\documentclass[10pt,twocolumn,letterpaper]{article}

\usepackage[pagenumbers]{cvpr} 

\definecolor{cvprblue}{rgb}{0.21,0.49,0.74}
\usepackage[pagebackref,breaklinks,colorlinks,allcolors=cvprblue]{hyperref}
\usepackage{multirow}
\usepackage{float}
\newcommand*\rot{\rotatebox{75 }}
\def\paperID{7624} 
\def\confName{CVPR}
\def\confYear{2026}

\title{When Semantics Regulate: Rethinking Patch Shuffle and Internal Bias for Generated Image Detection with CLIP}

\author{
Beilin Chu\textsuperscript{1},
Weike You\textsuperscript{1},
Mengtao Li\textsuperscript{1},
Tingting Zheng\textsuperscript{1},
Kehan Zhao\textsuperscript{1},\\
Xuan Xu\textsuperscript{1},
Zhigao Lu\textsuperscript{1},
Jia Song\textsuperscript{1},
Moxuan Xu\textsuperscript{2},
Linna Zhou\textsuperscript{1}\thanks{Corresponding author.} \\
\textsuperscript{1}School of CyberSpace Security, Beijing University of Posts and Telecommunications\\
\textsuperscript{2}	School of Finance, Central University of Finance and Economics\\
{\tt\small beilin.chu@bupt.edu.cn}
}

\begin{document}
\maketitle
\begin{abstract}
The rapid progress of GANs and Diffusion Models poses new challenges for detecting AI-generated images. Although CLIP-based detectors exhibit promising generalization, they often rely on semantic cues rather than generator artifacts, leading to brittle performance under distribution shifts. In this work, we revisit the nature of semantic bias and uncover that Patch Shuffle provides an unusually strong benefit for CLIP, that disrupts global semantic continuity while preserving local artifact cues, which reduces semantic entropy and homogenizes feature distributions between natural and synthetic images. Through a detailed layer-wise analysis, we further show that CLIP’s deep semantic structure functions as a regulator that stabilizes cross-domain representations once semantic bias is suppressed. Guided by these findings, we propose \textbf{SemAnti}, a semantic-antagonistic fine-tuning paradigm that freezes the semantic subspace and adapts only artifact-sensitive layers under shuffled semantics. Despite its simplicity, SemAnti achieves state-of-the-art cross-domain generalization on AIGCDetectBenchmark and GenImage, demonstrating that regulating semantics is key to unlocking CLIP’s full potential for robust AI-generated image detection.
\end{abstract}    
\section{Introduction}
\label{sec:intro}

Recent advances in large-scale generative models have blurred the boundary between authentic and synthesized visual content. Techniques such as diffusion models (e.g., Stable Diffusion \cite{rombach2022high}, DALL·E 2 \cite{ramesh2022hierarchical}) and generative adversarial networks (GANs) \cite{goodfellow2014generative} can now produce photo-realistic images with rich semantics and fine-grained textures, posing new challenges for multimedia forensics and trustworthy AI. The ability to automatically detect AI-generated images (AGIs) has thus become a crucial component of digital content authentication and AI safety.
\begin{figure}[t]
    \centering
    \includegraphics[width=\linewidth]{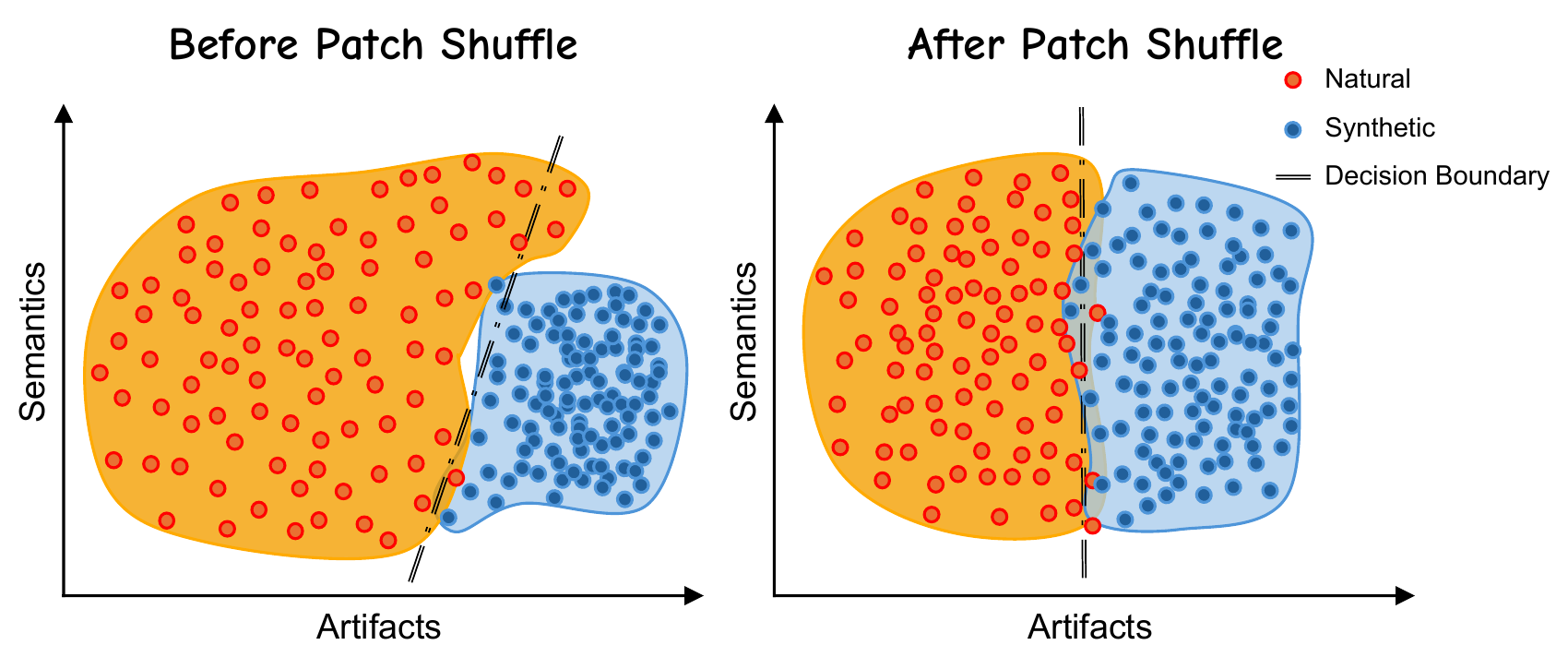}
    \caption{\textbf{Patch Shuffle} (PS) alleviates the semantic imbalance between natural and synthetic images. Before PS (left), natural samples exhibit richer and more diverse semantics, forming a \textbf{wide and dispersed} cluster, whereas synthetic samples remain \textbf{concentrated}, leading to a biased separation. After PS (right), PS \textbf{homogenizes} the distributions of both classes, enabling a more balanced and artifact-oriented decision boundary.}
    \label{figure:abstract}
\end{figure}

A series of artifact-based \cite{wang2020cnn,liu2022detecting} and frequency-based \cite{frank2020leveraging,qian2020thinking} approaches have established a solid foundation for AI-generated image detection (AGID). These methods exploit low-level statistical cues introduced during the generative process, including up-sampling artifacts, distinctive frequency patterns, and local texture inconsistencies. Their success demonstrates that generative models leave measurable and often universal traces that can be effectively captured by tailored feature extractors.

Despite rapid progress in AGID, a persistent and insufficiently understood issue remains. On unseen domains or semantic distributions, existing detectors experience severe accuracy degradation yet still achieve unexpectedly high Average Precision. This discrepancy indicates that detectors can still distinguish natural from synthetic images, but their decision boundaries become strongly biased under distribution shift. Prior work \cite{yan2024effort} attributes this problem to the asymmetric semantic coverage in AGID datasets. Natural images span diverse scenes, whereas synthetic images cluster around a few generative modes. As shown in Figure~1~(a), this imbalance encourages detectors to overfit narrow forgery patterns and rely on semantic shortcuts, causing cross-domain samples with richer semantics to be pushed toward the “natural” side of the boundary. Zheng et al. \cite{zheng2024breaking} offer another explanation by showing that AGIs contain two types of signals: generator artifacts, which arise from generator-specific upsampling, and semantic artifacts, which are scene- or dataset-specific patterns correlated with semantics. Generator artifacts cause overfitting across generators, while semantic artifacts dominate failure under semantic shifts. For example, a detector trained on “church” images struggles on “bedroom.” Together, these \textbf{two findings} indicate that avoiding dataset-specific semantic fitting and suppressing semantic artifacts are crucial for generalizable AGI detection.

This perspective naturally motivates the use of CLIP \cite{radford2021learning}. Through large-scale vision–language pretraining on diverse natural images, CLIP learns a stable and highly structured semantic manifold. Images with similar semantics cluster close to one another, and different semantic categories remain clearly separated. Such a representation space appears ideal for providing a semantically structured manifold that can help the detector resist overfitting to semantic artifacts. This semantic regularity should, in principle, guide the detector toward learning generator artifacts instead of memorizing dataset-specific semantics. However, experiments show that directly fine-tuning CLIP still results in considerable degradation under cross-domain evaluation. This observation suggests that, despite strong semantic priors, CLIP remains vulnerable to the semantic artifacts inherited from the training data.

A striking observation emerges when we apply Patch Shuffle (PS) to images before fine-tuning. By randomly permuting patch order while keeping the patch content intact, PS produces a dramatic improvement in cross-domain generalization. This improvement cannot be fully explained by conventional augmentation arguments. Additional experiments (see Section~\ref{sec:visal}) reveal that, after training with PS, CLIP’s latent space becomes more consistent across domains. Semantic clusters remain coherent, yet within each cluster, natural and synthetic samples remain cleanly separable. This suggests that PS guides CLIP away from spurious scene specific semantics while preserving its robust semantic structure. Under this configuration, CLIP naturally performs what we call semantic-consistent artifact learning, where semantics provide a stable embedding scaffold and generator artifacts become the main discriminative signal within each semantic cluster.

These insights prompt us to further ask how such a seemingly simple operation can yield such nontrivial generalization effects: \textbf{(RQ1)} \textit{Does PS universally benefit diverse architectures, or is its effectiveness unique to CLIP’s vision–language pretraining?} \textbf{(RQ2)} \textit{What representation shifts occur within CLIP after applying PS, and how are these shifts associated with better cross-domain robustness?} Motivated by these questions, we conduct an in-depth investigation into how PS interacts with the training dynamics of CLIP. Our analysis reveals that PS simultaneously addresses the generalization challenges identified in the \textbf{two findings} above. First, by disrupting global spatial coherence, PS effectively lowers the semantic entropy of training images and homogenizes the distribution of natural and synthetic samples, thereby restoring semantic balance during training (see Figure~\ref{figure:abstract} right). Second, PS substantially weakens CLIP’s reliance on semantic artifacts, mitigating the overfitting induced by scene-specific patterns. Notably, we observe that this benefit is largely unique to CLIP, suggesting that CLIP’s well-formed semantic space serves as a natural regulator of cross-domain generalization, constraining the detector’s decision boundaries to remain aligned with semantics that are stable across domains while allowing artifact cues to become the dominant source of discriminative evidence. A comprehensive layer-wise analysis further shows that the improvements predominantly arise from the deeper blocks, whose representations encode richer and more transferable semantic structure. To this end, we introduce SemAnti, a CLIP-based training paradigm that combines PS with selective layer freezing to preserve indispensable semantic priors while steering the model toward artifact-oriented, domain-robust representation learning.

Our contributions are threefold:
\begin{itemize}
	\item We revisit the significance of PS in CLIP-based detectors and empirically demonstrate its remarkable ability to enhance generalization. The observed gain appears unique to CLIP’s large-scale vision–language pretraining, motivating a deeper exploration of its internal mechanism.
	\item We are the first to systematically examine the internal representational behavior of CLIP-based detectors across different network depths. Our analysis shows that CLIP’s deeper layers, endowed with a highly structured semantic manifold, act as an effective regulator that suppresses semantic artifacts and plays an indispensable role in achieving strong cross-domain robustness.
	\item Building upon these findings, we propose \textbf{SemAnti}, a CLIP-based training paradigm tailored for semantics-consistent generalizable detection. Extensive experiments confirm our hypotheses and further demonstrate the superior generalization performance.
\end{itemize}
\begin{figure*}[ht]
    \centering
    \includegraphics[width=0.9\linewidth]{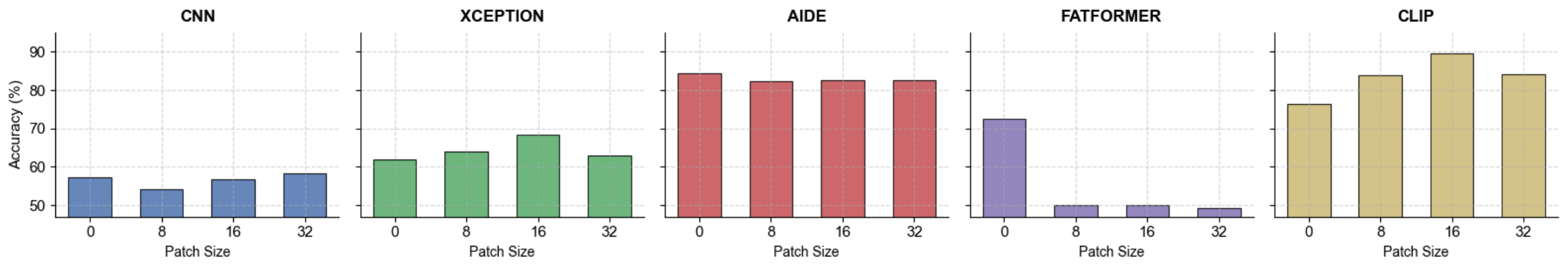}
    \caption{Accuracy of five detectors under different Patch Shuffle (PS) patch sizes (PS=0 denotes no shuffle). Only CLIP shows significant improvement, indicating that PS effectively mitigates its reliance on semantics and enhances artifact-level generalization.}
    \label{figure:intro_1}
\end{figure*}
\section{Related Works}
\subsection{CLIP-based Foundation Detectors}
The advent of large vision-language models, such as CLIP \cite{radford2021learning}, has fundamentally reshaped AIGC detection research. CLIP’s contrastive pretraining aligns images and text across massive web-scale data. This makes CLIP an appealing backbone for detecting AI-generated content due to its inherent robustness and transferability.
Ojha et al. \cite{ojha2023towards} first showed that a frozen CLIP image encoder with a linear classifier can already achieve strong cross-generator detection performance. Follow-up works explored adapting CLIP through fine-tuning or prompt learning. Sha et al. \cite{sha2023fake} introduced DE-FAKE, which injects textual prompts into CLIP to enhance attribution and detection, while Yan et al. \cite{yan2024sanity} proposed AIDE, a CLIP-based mixture-of-experts framework that combines DCT-selected local artifact cues with high-level semantics. Despite these advances, recent studies report that fine-tuning can distort CLIP’s semantic manifold, resulting in reduced robustness under scene- or texture-level distribution shifts \cite{yan2024effort}. Co-SPY \cite{co-spy} detects synthetic images by fusing CLIP-based semantic features with pixel-level artifacts, and dynamically adjusts the contribution of semantics and local cues based on the input content.

\subsection{Local Forensic Cues Beyond Semantics}
Beyond designing task-specific architectures, another direction seeks to improve generalization through patch-level or texture-aware representations. Chai et al. \cite{chai2020makes} reveal that local patch classifiers generalize better across unseen generators, as small receptive fields emphasize pixel-level inconsistencies rather than high-level semantics.
A series of recent studies \cite{wang2023dire,cazenavette2024fakeinversion,FIRE} detect diffusion-generated images by reconstructing them through diffusion models and measuring the reconstruction error as a reliability cue for discrimination. PatchCraft defines a custom metric to measure patch texture richness, then reorders image patches by this score to amplify texture contrast for better performance. Yang et al. \cite{D3} proposes a parallel branch processes patch shuffled images as auxiliary discrepancy signals to capture universal generator artifacts.
Yan et al. \cite{yan2024sanity} extracts low-level patch statistics by computing DCT scores for all image patches, selecting those with the highest and lowest frequencies, and feeding them into SRM convolutional blocks.
Unlike existing patch-based approaches that often rely on specialized architectures or hand-crafted pre-processing (e.g., texture ranking, high-pass filters), our work finds that simply introducing patch shuffling during the training of CLIP-based detectors can markedly enhance their generalization capability. We further conduct a series of analyses to interpret this phenomenon and uncover the underlying mechanisms that drive such improvements.
\section{Methodology}
\begin{figure*}[ht]
    \centering
    \includegraphics[width=0.8\linewidth]{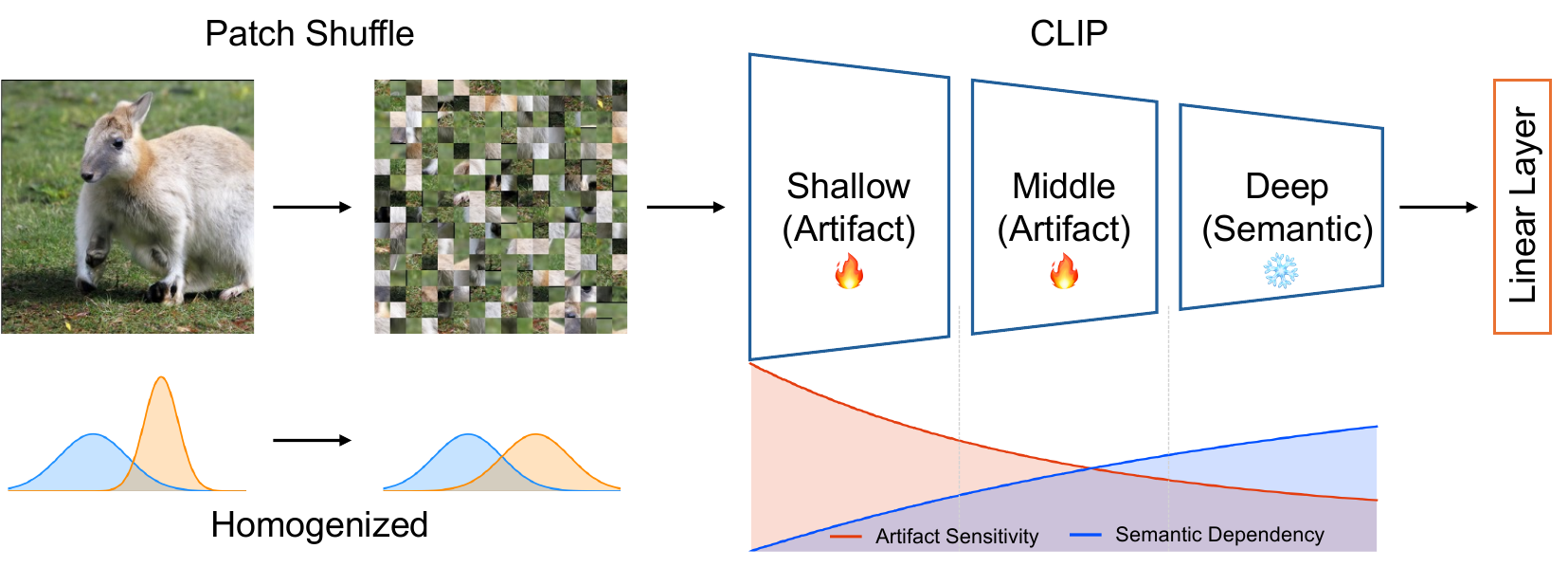}
    \caption{Overview of our architecture. Patch Shuffle homogenizes image semantics, balancing the representation distributions of natural and synthetic samples and steering the model toward local artifact cues. In addition, we freeze the semantic-rich deep layers of CLIP to preserve high-level semantic artifact priors, enabling more robust cross-domain generalization.}
    \label{figure:archi}
\end{figure*}
\subsection{Analysis}
In this section, we first address \textbf{RQ1} and investigate whether PS can universally improve the generalization ability of different network architectures. Specifically, we conduct experiments on the SDv4 subset of GenImage by training detectors based on convolutional (CNN \cite{wang2020cnn}, Xception \cite{chollet2017xception}) and transformer architectures (AIDE \cite{yan2024sanity} and Fatformer \cite{liu2024forgery}), and compare them with a CLIP visual encoder equipped with LoRA fine-tuning. As shown in Figure~\ref{figure:intro_1}, unlike CLIP, the other models exhibit performance degradation or even fail to converge when trained on patch-shuffled images.

These results indicate that the gain from PS is not a generic property shared across architectures. Instead, it is closely related to CLIP’s vision–language pretraining. For CNNs and ViTs that are trained from scratch, semantics and low-level statistics are intertwined within a single task-specific feature space. Once patches are randomly permuted, the global layout information that these models depend on for learning both scene semantics and generator-related cues is severely disrupted. As a result, training tends to collapse or drift toward noisy patterns.

In contrast, CLIP begins with a pretrained embedding space in which global semantic relationships have already been organized by large-scale language supervision. This semantic structure forms a stable manifold that remains robust even when local spatial coherence is perturbed. Under such a prior, PS selectively breaks scene-level spatial patterns and reduces the semantic entropy that varies across classes, while leaving local textural and statistical irregularities largely unaffected. This process suppresses semantic artifacts that originate from specific scenes or datasets and prevents the model from relying on these spurious cues.

\subsection{SemAnti}
To further understand how CLIP internally learns semantics and artifacts, we conduct a comprehensive layer-wise analysis, as detailed in Section~\ref{sec:layer_analysis}.
Our observations reveal that different layers exhibit distinct tendencies toward artifact and semantic representations.
Based on this analysis, we divide the parameters of CLIP into two functional groups.
The first group consists of layers that are artifact-specialized, capable of capturing low-level inconsistencies and learning task-relevant features that directly contribute to forgery discrimination.
The second group serves as a high-level semantic knowledge base, which by itself cannot detect generalizable artifacts but plays an indispensable role in maintaining representation stability and preventing overfitting to the training semantics.
This semantic component thus functions as an inherent regularizer, stabilizing optimization and preserving CLIP’s pretrained generalization prior.

Building upon these findings, we introduce the \textbf{SemAnti} (\textbf{Sem}antic-\textbf{An}tagonistic Fine-\textbf{t}un\textbf{i}ng) training paradigm.
Specifically, SemAnti freezes the deep semantic-dominant blocks to preserve CLIP’s stable semantic structure, while fine-tuning only the artifact-sensitive early and middle blocks under Patch-Shuffled inputs. This prevents semantic re-dominance during adaptation and strengthens the model’s reliance on transferable artifact cues rather than scene-dependent semantics.
Let the parameters of CLIP be represented as $\theta = \{\theta_A, \theta_S\}$, where $\theta_A$ and $\theta_S$ denote the artifact-specialized and semantic parameters, respectively.
Given a training image x with label $y \in \{0, 1\}$ (natural or generated), our optimization objective can be written as:
\begin{align}
    \mathcal{L}(g(F(x; \theta_A, \theta_S)), y),\tag{1}
\end{align}
where $F(\cdot)$ represents the CLIP-based image encoder, $g(\cdot)$ denotes a linear classifier.

During optimization, only the parameters in the artifact-related subspace are updated, while the semantic parameters remain frozen.
The gradient updates follow:
\begin{align}
\theta_A \leftarrow \theta_A - \alpha \frac{\partial \mathcal{L}(g(F(x)),\,y)}{\partial \theta_A},
\qquad
\theta_S \leftarrow \theta_S,
\tag{2}
\end{align}
where $\alpha$ is the learning rate.
This selective optimization ensures that the model adapts to generator-specific statistics without disturbing the pretrained semantic prior.

By decoupling CLIP’s semantic manifold from the adaptation process, SemAnti suppresses semantic artifacts and compels the model to anchor its decisions on generator artifact, while preserving the semantic structure that regularizes and balances the overall classification space.
\begin{figure*}[ht]
    \centering
    \includegraphics[width=\linewidth]{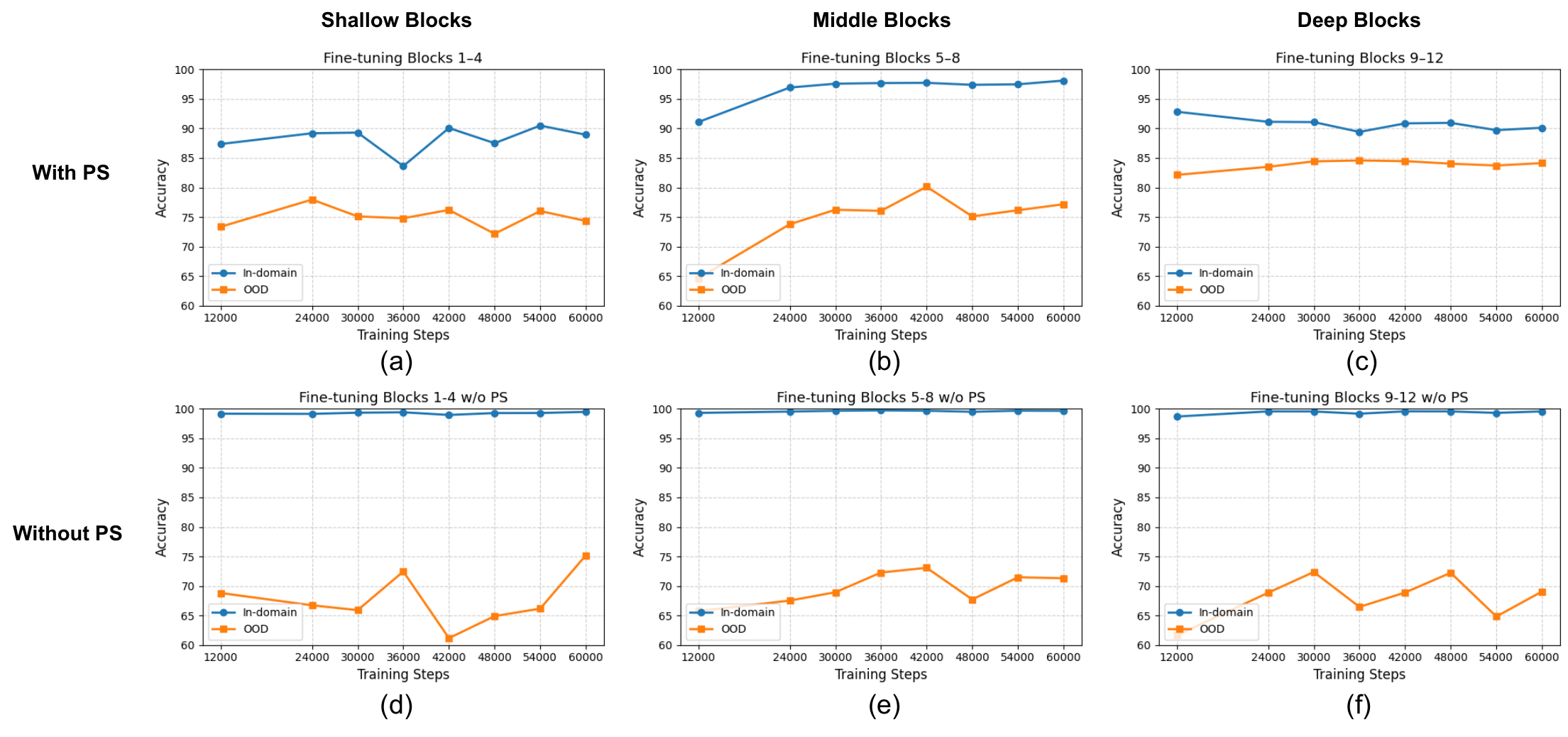}
    \caption{Layer-wise fine-tuning analysis of CLIP. 
    Fine-tuning shallow layers enhances artifact sensitivity, 
    while deeper layers lead to overfitting to semantic cues.}
    \label{figure:layer_analysis}
\end{figure*}
\section{Experiments}
\subsection{Set up}
\textbf{Dataset.} We conduct experiments on two large-scale benchmarks that comprehensively evaluate the generalization ability of AGI detectors.
GenImage \cite{zhu2024genimage} is a million-scale dataset covering a broad range of generative models, including 7 diffusion models (ADM \cite{dhariwal2021diffusion}, GLIDE \cite{intro_308}, Midjourney \cite{midjourney}, Stable Diffusion v1.4 \cite{StableDiffusion}, Stable Diffusion v1.5 \cite{StableDiffusion}, VQDM \cite{intro_312}, and WuKong \cite{wukong}) and 1 GAN model (BigGAN \cite{intro_302}). Each subset contains paired real and synthetic images under identical prompts.
AIGCDetectBenchmark \cite{exper_RPTC} is built upon GenImage and further expands its coverage of generative models.
On the GAN side, it adds ProGAN \cite{karras2017progressive}, StyleGAN \cite{karras2019style}, StyleGAN2 \cite{intro_306}, CycleGAN \cite{intro_303}, StarGAN \cite{intro_304}, GauGAN \cite{intro_305}, and WFIR (WhichFaceIsReal) \cite{WFIR}.
On the Diffusion side, it additionally introduces DALL·E 2 \cite{ramesh2022hierarchical}. In total, it comprises 16 representative generative models, covering both GAN-based and diffusion-based architectures, making it suitable for assessing cross-generator and cross-domain robustness.

\noindent\textbf{Implementation Details.} During training and testing, each image is center-cropped to 224 $\times$ 224 without any resizing. If either side of the original image is smaller than 224 pixels, we replicate the image along both spatial axes until its dimensions exceed 224 $\times$ 224, and then apply a central crop.
This procedure avoids any resampling operations such as bilinear interpolation, which would introduce additional up-sampling or down-sampling artifacts. We found that these resizing-induced distortions can easily interfere with generator-specific cues and significantly degrade cross-model generalization. Related experiment is detailed in the appendix.
To ensure the reproducibility of our findings, we train two variants of CLIP, ViT-L/14 and ViT-B/32, to demonstrate the generality of our approach.
During fine-tuning, we adopt the LoRA technique to efficiently adapt the pretrained CLIP weights while keeping most parameters frozen.
Training is performed with the Adam optimizer, an initial learning rate of 4e-4, and a batch size of 32. All experiments are conducted on two NVIDIA RTX 4090 GPU.

\noindent\textbf{Metrics.} For evaluation, we follow prior works \cite{yan2024sanity, wang2020cnn} and adopt two widely used metrics: Accuracy (Acc) and Average Precision (AP).

\subsection{Layer-wise Analysis}
\label{sec:layer_analysis}
To answer \textbf{RQ2} and clarify how PS changes CLIP’s reliance on semantics vs artifact cues, we conduct a layer-wise analysis to identify what each block contributes. We freeze the backbone parameters of ViT-B/32 and selectively fine-tune specific blocks at different depths. In addition, we introduce PS to analyze its effect on network training and generalization. The experiments are conducted on the GenImage dataset, using the SDv4 subset for training and evaluating performance across eight test subsets. To directly examine the impact of different settings on model generalization, subsets with similar distributions—namely SDv4, SDv5, and Wukong—are grouped as in-domain, while the remaining subsets are regarded as out-of-domain (OOD).

As shown in Figure~\ref{figure:layer_analysis} (d–f), when PS is not applied, all three fine-tuning depths exhibit a similar behavior: the model rapidly overfits the in-domain data, reaching nearly 100\% accuracy, whereas OOD accuracy fluctuates around 65\%. This pattern indicates that, without PS, the CLIP backbone tends to memorize the global semantic structures of the training data rather than focusing on intrinsic generative artifacts, leading to severe semantic bias and poor cross-domain generalization.
\begin{table*}[ht]
\resizebox{\linewidth
}{!}
{%
\begin{tabular}{lccccccccccccccccc}
\toprule
Method&\rot{ProGAN} &\rot{StyleGAN} & \rot{BigGAN} & \rot{CycleGAN} &\rot{StarGAN} &\rot{GauGAN} &\rot{StyleGAN2} &\rot{WFIR} &\rot{ADM} &\rot{Glide} & \rot{{Midjourney}} & \rot{{SD v1.4}} & \rot{{SD v1.5}}& \rot{{VQDM}}& \rot{{Wukong}}& \rot{{DALLE2}} & \rot{\textit{Mean}} \\ \midrule
CNNSpot \citep{wang2020cnn}   & \textbf{100.00} & 99.83          & 85.99          & 94.94           & 99.04           & 90.82           & 99.48          & \textbf{99.85} & 75.67          & 72.28          & 66.24          & 61.20          & 61.56          & 68.83          & 57.34          & 53.51          & 80.41          \\
FreDect \citep{frank2020leveraging}   & \underline{99.99}     & 88.98          & 93.62          & 84.78           & 99.49           & 82.84           & 82.54          & 55.85          & 61.77          & 52.92          & 46.09          & 37.83          & 37.76          & 85.10          & 39.58          & 38.20          & 67.96          \\
Fusing \citep{ju2022fusing}    & \textbf{100.00} & 99.50          & 90.70          & 95.50           & 99.80           & 88.30           & 99.60          & 93.30          & 94.10          & 77.50          & 70.00          & 65.40          & 65.70          & 75.60          & 64.60          & 68.12          & 84.23          \\
LNP \citep{liu2022detecting}       & 99.89           & 98.60          & 84.32          & 92.83           & \textbf{100.00} & 78.85           & 99.59          & 91.45          & 94.20          & 88.86          & 76.86          & 94.31          & 93.92          & 87.35          & 92.38          & 96.14          & 91.85          \\
LGrad \citep{tan2023learning}     & \textbf{100.00} & 98.31          & 92.93          & 95.01           & \textbf{100.00} & 95.43           & 97.89          & 57.99          & 72.95          & 80.42          & 71.86          & 62.37          & 62.85          & 77.47          & 62.48          & 82.55          & 81.91          \\
UnivFD \cite{ojha2023towards}    & 99.08           & 91.74          & 75.25          & 80.56           & 99.34           & 72.15           & 88.29          & 60.13          & 85.84          & 78.35          & 61.86          & 49.87          & 49.52          & 54.57          & 55.38          & 74.48          & 73.53          \\
DIRE \citep{wang2023dire}    & 58.79           & 56.68          & 46.91          & 50.03           & 40.64           & 47.34           & 58.03          & 59.02          & \textbf{99.79} & \textbf{99.54} & \textbf{97.32} & 98.61          & 98.83          & 98.98          & 98.37          & \textbf{99.71} & 75.54          \\
NPR \citep{tan2024rethinking}       & \textbf{100.00} & 99.81          & 87.87          & 98.55           & 99.90           & 85.57           & 99.90          & 65.38          & 74.61          & 85.73          & 85.41          & 84.02          & 84.67          & 81.20          & 80.51          & 76.72          & 86.87          \\
FatFormer \citep{liu2024forgery} & \textbf{100.00} & 99.75          & \underline{99.98}    & \textbf{100.00} & \textbf{100.00} & \textbf{100.00} & \underline{99.92}    & 98.48          & 91.73          & 95.99          & 62.76          & 81.12          & 81.09          & 96.99          & 85.86          & 81.84          & 92.22          \\
DRCT \citep{drct}      & 91.03           & 79.44          & 93.51          & 98.68           & 96.29           & 86.62           & 73.80          & 91.04          & 88.96          & 94.64          & \underline{97.03}    & \underline{99.65}    & \underline{99.49}    & 96.54          & \underline{99.37}    & 97.67          & 92.74          \\
D³ \citep{D3}        & \textbf{100.00} & 97.59          & 98.18          & \underline{99.91}     & 98.84           & 99.04           & 96.52          & 95.78          & 95.86          & 92.42          & 80.15          & 84.83          & 84.96          & 93.38          & 85.44          & 81.34          & 92.77          \\
CO-SPY \citep{co-spy}    & \textbf{100.00} & 98.35          & 97.11          & 99.74           & 98.73           & 99.21           & 98.85          & 93.98          & 90.50          & 96.39          & 88.42          & 92.86          & 92.87          & 96.28          & 93.72          & 93.54          & 95.67          \\
PatchCraft \citep{exper_RPTC} & \textbf{100.00} & 98.96          & 99.42          & 85.26           & \textbf{100.00} & 81.33           & 97.74          & 95.26          & 93.40          & 94.04          & 96.48          & 99.06          & 99.06          & 96.26          & 97.54          & \underline{99.56}    & 95.84          \\
AIDE \citep{yan2024sanity}      & \textbf{100.00} & \textbf{99.99} & 94.44          & 99.89           & \underline{99.99}     & 97.69           & \textbf{99.96} & \underline{99.27}    & \underline{98.77}    & \underline{98.94}    & 88.13          & 98.26          & 98.20          & \underline{99.27}    & 98.62          & 99.41          & \underline{98.18}    \\
\hline
Ours       & \textbf{100.00} & \underline{99.97}    & \textbf{99.99} & \textbf{100.00} & \textbf{100.00} & \underline{99.99}     & \textbf{99.96} & 94.37          & 98.21          & 98.82          & 95.57          & \textbf{99.80} & \textbf{99.76} & \textbf{99.57} & \textbf{99.41} & 98.95          & \textbf{99.02}\\
\bottomrule					
\end{tabular}
}
\caption{\textbf{ Comparison on the AIGCDetectBenchmark \citep{exper_RPTC}.} Average precision (AP \%) of different detectors (rows) in detecting natural and synthetic images from different generators (columns). These methods are trained on ProGAN and evaluated over sixteen generators. The best result and the second-best result are marked in \textbf{bold} and \underline{underline}, respectively. Full results are provided in the appendix}
\label{table:b1_ap_appendix}
\end{table*}
\begin{table*}[ht]
\centering
\resizebox{0.9\linewidth}{!}
{\small
\begin{tabular}{lccccccccc}
\hline
Method & Midjourney & SD v1.4 & SD v1.5 & ADM & GLIDE & Wukong & VQDM & BigGAN & \textit{Mean} \\ \hline
{ResNet-50} \citep{he2016deep} & 54.90 & \textbf{99.90} & 99.70 & 53.50 & 61.90 & 98.20 & 56.60 & 52.00 & 72.09 \\
{DeiT-S} \citep{touvron2021training} & 55.60 & \textbf{99.90} & 99.80 & 49.80 & 58.10 & 98.90 & 56.90 & 53.50 & 71.56 \\
{Swin-T} \citep{liu2021swin} & 62.10 & \textbf{99.90} & 99.80 & 49.80 & 67.60 & 99.10 & 62.30 & 57.60 & 74.78 \\
{CNNSpot} \citep{wang2020cnn} & 52.80 & 96.30 & 95.90 & 50.10 & 39.80 & 78.60 & 53.40 & 46.80 & 64.21 \\
{Spec} \citep{zhang2019detecting} & 52.00 & 99.40 & 99.20 & 49.70 & 49.80 & 94.80 & 55.60 & 49.80 & 68.79 \\
{F3Net} \citep{qian2020thinking} & 50.10 & \textbf{99.90} & \textbf{99.90} & 49.90 & 50.00 & \textbf{99.90} & 49.90 & 49.90 & 68.69 \\
{GramNet} \citep{liu2020global} & 54.20 & 99.20 & 99.10 & 50.30 & 54.60 & 98.90 & 50.80 & 51.70 & 69.85 \\
{DIRE} \citep{wang2023dire} & 60.20 & \textbf{99.90} & \underline{99.80} & 50.90 & 55.00 & \underline{99.20} & 50.10 & 50.20 & 70.66 \\
{UnivFD} \citep{ojha2023towards} & 73.20 & 84.20 & 84.00 & 55.20 & 76.90 & 75.60 & 56.90 & \underline{80.30} & 73.29 \\
{GenDet} \citep{zhu2023gendet} & \underline{89.60} & 96.10 & 96.10 & 58.00 & 78.40 & 92.80 & 66.50 & 75.00 & 81.56 \\
{PatchCraft} \citep{zhong2023rich} & 79.00 & 89.50 & 89.30 & 77.30 & 78.40 & 89.30 & 83.70 & 72.40 & 82.30 \\
{DRCT} \citep{drct} & 82.39 & 88.45 & 88.44 & 76.81 & 86.60 & 87.75 & \underline{84.07} & 74.10 & 83.58 \\
{CO-SPY} \citep{co-spy} & 83.45 & 96.83 & 96.68 & 67.25 & \textbf{93.02} & 95.93 & 78.83 & 65.20 & 84.65 \\
{AIDE} \citep{yan2024sanity} & 79.38 & \underline{99.74} & 99.76 & \underline{78.54} & \underline{91.82} & 98.65 & 80.26 & 66.89 & \underline{86.88} \\ \hline
\textbf{Ours} & \textbf{95.34} & 97.93 & 97.65 & \textbf{83.41} & 81.08 & 96.68 & \textbf{90.43} & \textbf{88.03} & \textbf{91.32} \\ \hline
\end{tabular}
}
\caption{\textbf{Comparison on the GenImage \citep{zhu2024genimage}}. Accuracy (\%) of different detectors (rows) in detecting natural and synthetic images from different generators (columns). These methods are trained on 
natural images from ImageNet and synthetic images generated by SD v1.4 and evaluated over eight generators. The best result and the second-best result are marked in \textbf{bold} and \underline{underline}, respectively.}
\label{table:b2}
\end{table*}

After introducing PS (Figure~\ref{figure:layer_analysis} (a–c)), the results change markedly. The improvement in the shallow and middle layers remains relatively modest (from 67\% to 75\%), because these layers primarily encode low-level appearance cues that are less influenced by semantic bias and therefore less affected by PS. In contrast, fine-tuning the deeper blocks yields a substantial performance gain (from 68\% to 84\%). These layers contain CLIP’s rich semantic representational structure, which normally dominates the decision process and is prone to absorbing semantic artifacts from the training domain. PS disrupts this dominance by weakening class- and scene-dependent semantic continuity while keeping local artifact traces intact. As a result, the deeper layers undergo a more pronounced redistribution of representational emphasis, shifting from semantic reliance toward artifact-oriented discrimination. This explains why PS produces its strongest effect precisely where semantic regulation is most needed.
\begin{figure*}[ht]
    \centering
    \includegraphics[width=0.97\linewidth]{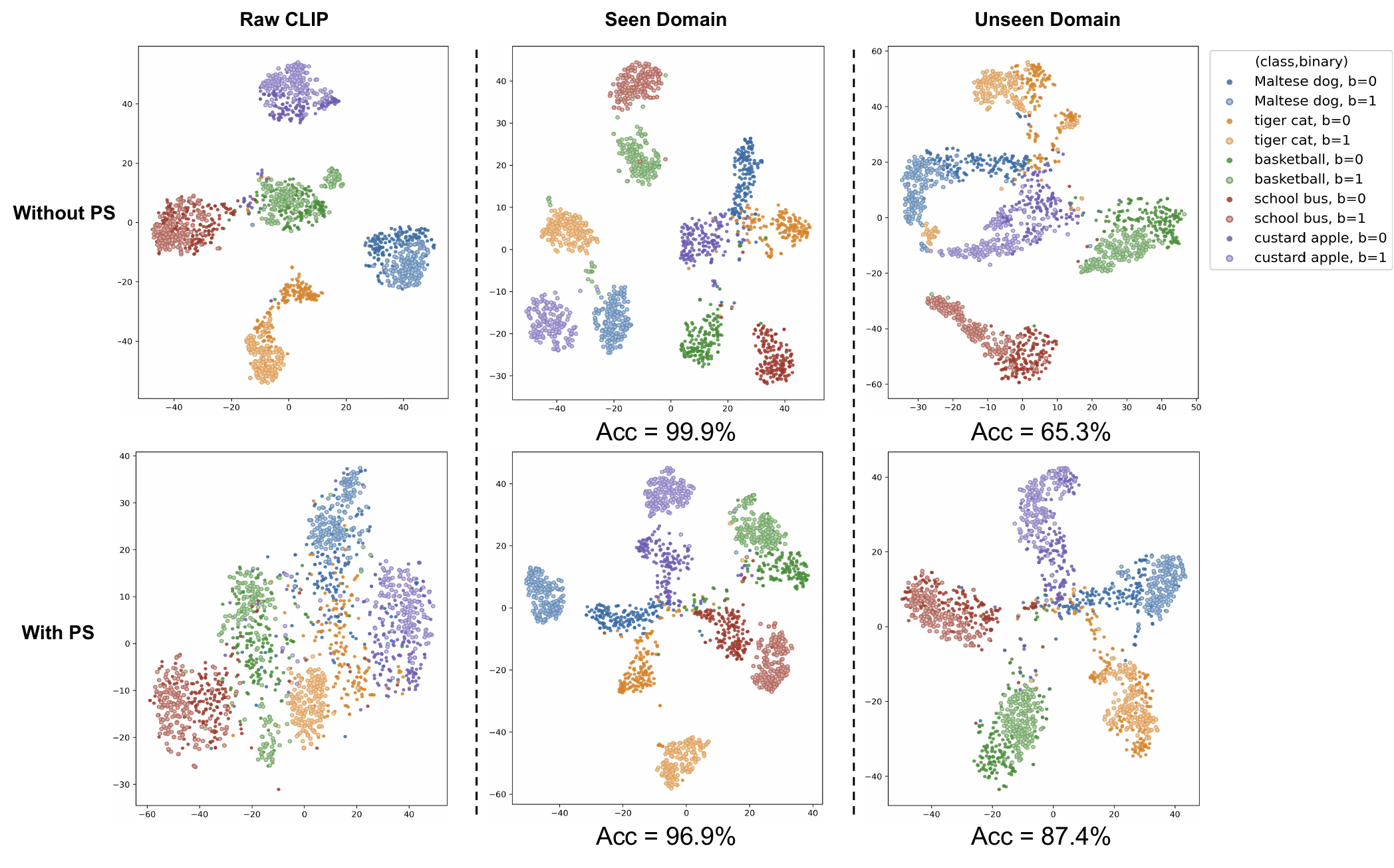}
    \caption{Visualization of the latent distributions of the original CLIP and our method under different settings. More visualizations are provided in the appendix.}
    \label{figure:visual}
\end{figure*}

Although fine-tuning the deeper layers with PS delivers the largest OOD improvement, this result also indicates that the deep semantic space is the main source of domain bias. CLIP’s deeper layers encode the high-level semantic structure formed during large-scale pretraining. Without proper constraints during fine-tuning, this semantic structure becomes entangled with scene-dependent semantic artifacts, which are subtle correlations between image content and generative traces. PS strongly suppresses this entanglement in the deeper layers, which explains why the most substantial gains appear there.
This insight suggests an important design principle. Since the deeper layers contain the semantic geometry that stabilizes representations across domains, they should be preserved rather than substantially altered. Therefore, our final model adopts an inverse training strategy: we freeze the deeper blocks to retain their pretrained semantic structure and fine-tune only the shallow and middle layers. This approach maintains CLIP’s transferable semantic scaffold while preventing the deeper layers from absorbing domain-specific semantic artifacts. As a result, the model learns a more balanced and robust representation that achieves strong cross-domain generalization in AGID.

\subsection{Comparision to State-of-the-art}
\textbf{On the AIGCDetectBenchmark:}
Here we use ViT-L/14 as visual backbone and report in AP. All methods are trained on ProGAN and tested on different subsets. As summarized in Table~\ref{table:b1_ap_appendix}, our results reveal several consistent trends across different generators.
Traditional local artifact–based detectors, such as CNNSpot, LGrad, and DIRE, achieve strong detection accuracy on GAN-series generators but experience a pronounced drop when transferred to diffusion-generated images, indicating limited cross-domain robustness.
Compared with these recent state-of-the-art methods, our model achieves the highest overall performance, surpassing AIDE by 0.84\%, PatchCraft by 3.18\%, CO-SPY by 3.35\%, and D$^3$ by 6.25\% in average AP.
Furthermore, our approach obtains the best results on several key subsets, including SDv4, SDv5, VQDM, and WuKong.
Considering that AIDE, CO-SPY, and D$^3$ are all pretrained on large-scale vision foundation models, the superior performance of our method indicates its stronger ability to guide CLIP to achieve better generalization.

\noindent\textbf{On the GenImage:}
To further validate the generalization capability of our method, we employ ViT-B/32 as the visual backbone and report Acc as the main evaluation metric. All methods are trained on SDv4 and tested across multiple subsets.
As shown in Table~\ref{table:b2}, our method consistently surpasses all state-of-the-art approaches, exceeding AIDE by 4.44\%, CO-SPY by 6.67\%, and DRCT by 7.74\% in average accuracy.
Notably, our model attains remarkable accuracies of 95.34\% and 88.03\% on the more challenging Midjourney and BigGAN subsets, where most existing methods perform close to random guessing.
Since the SDv4 training set contains higher-resolution and more diverse images than ProGAN, making it closer to real-world detection scenarios, the PatchCraft model, which lacks rich semantic priors, suffers a noticeable decline in generalization on this benchmark.
Although our approach does not lead on the training set SDv4 and its distributionally similar SDv5 subset, this behavior highlights that the proposed method effectively suppresses overfitting to semantic artifacts and focuses on learning artifact signals that generalize across diverse distributions.

\subsection{Visualization}
\label{sec:visal}
To visually illustrate the effectiveness of our method in balancing generator and semantic artifacts, we employ t-SNE to visualize the latent distributions of natural and synthetic images under different settings. As shown in Figure~\ref{figure:visual}, without PS, CLIP strongly clusters images by semantics, showing clear semantic grouping. After applying PS, these semantics-driven clusters are largely broken, and the feature space becomes more balanced, indicating that PS effectively suppresses semantic dominance and prevents CLIP from overfitting to semantic bias.

For SemAnti on the in-domain SDv4 test set, training without PS yields a clean natural–synthetic split but destroys semantic coupling: images with similar semantics no longer stay close. Although this model reaches nearly 100\% accuracy, the latent space becomes semantically distorted. With PS, SemAnti achieves a still-strong 97\% accuracy while preserving both structural dimensions: natural samples gather near the center, synthetic samples spread toward the periphery, and semantic neighbors remain close. The latent space becomes well organized along both the semantic and authenticity axes.

On the OOD ADM set, SemAnti without PS produces entangled embeddings that mix categories such as cat, dog, and bus, and fail to form a stable natural–synthetic boundary. This collapse is reflected in its low 65\% accuracy. In contrast, SemAnti with PS maintains a well-structured distribution, showing clear semantic clusters and consistent natural–synthetic separation, and achieves a remarkable 87.4\% accuracy. These results confirm that PS breaks CLIP’s global semantic dependence and enables SemAnti to leverage semantic-consistent artifact cues that generalize robustly across domains.

\subsection{Ablation Studies}
\noindent\textbf{Effect of Patch Shuffle Size.} We further investigate the influence of the shuffle granularity on the detection performance by varying the patch size from 8 to 32 for two CLIP variants, ViT-L/14 and ViT-B/32. As shown in Table~\ref{table:ablation}, the optimal performance is achieved when the shuffle size is set to 16 for ViT-L/14 and 14 for ViT-B/32, reaching 90.39\% and 85.73\% in mean accuracy, respectively. We attribute this phenomenon to the balance between disrupting global semantics and preserving local artifact consistency. Smaller shuffle sizes (e.g., 8) fail to sufficiently remove semantic correlations, while excessively large sizes (e.g., 32) destroy fine-grained artifacts. Moderate shuffling (14–16) effectively homogenizes semantics across domains while maintaining distinguishable local cues, thereby enhancing the model’s cross-generator generalization.

\noindent\textbf{Effect of Patch Shuffle and Layer Freezing.} We further perform an ablation study to evaluate the contribution of the Patch Shuffle (PS) mechanism and the layer-freezing strategy. As shown in Table~\ref{table:ablation2}, applying PS alone brings a remarkable improvement (+8.4\% in Acc), confirming that disrupting global semantics effectively steers the model toward artifact-oriented cues. Building upon PS, we explore different freezing depths of the CLIP visual transformer. Freezing the early layers (1–4) severely hurts performance due to the loss of low-level texture adaptability, while progressively freezing deeper layers yields a consistent improvement. The best performance is achieved when freezing the last 8–12 blocks (90.12\% / 96.73\%), indicating that preserving learnable shallow representations while keeping the high-level semantic structure fixed offers the best balance between semantic invariance and artifact discrimination.

\begin{table}[ht]
\centering
\resizebox{0.85\columnwidth}{!}{
\begin{tabular}{cccc|c}
\hline
\multicolumn{1}{l}{Model} & \multicolumn{1}{l}{PatchSize} & AIGC           & Genimage       & \textit{Mean}  \\ \hline
\multirow{4}{*}{Vit-L/14} & \multicolumn{1}{c|}{8}        & 78.73          & 83.92          & 81.32          \\
                          & \multicolumn{1}{c|}{14}       & 89.93          & 82.58          & 86.26          \\
                          & \multicolumn{1}{c|}{16}       & \textbf{94.90} & 85.88          & \textbf{90.39} \\
                          & \multicolumn{1}{c|}{32}       & 80.88          & 79.45          & 80.17          \\ \hline
\multirow{4}{*}{Vit-B/32} & \multicolumn{1}{c|}{8}        & 76.78          & 83.87          & 80.32          \\
                          & \multicolumn{1}{c|}{14}       & 81.81          & \textbf{89.66} & \textbf{85.73} \\
                          & \multicolumn{1}{c|}{16}       & 71.22          & 84.15          & 77.69          \\
                          & \multicolumn{1}{c|}{32}       & 83.99          & 84.05          & 84.02          \\ \hline
\end{tabular}}
\caption{\textbf{Ablation on the patch shuffle size.}
Detection accuracy (\%) on AIGCDetectBenchmark and GenImage using two CLIP-based backbones under different shuffle granularities. The best results for each backbone are highlighted in bold. More ablation studies are provided in the appendix.}
\label{table:ablation}
\end{table}
\begin{table}[ht]
\centering
\begin{tabular}{cccc|cc}
\hline
Setting                                                                                                        &                           &                           &                           & Acc   & AP    \\ \hline
\multicolumn{1}{c|}{\multirow{2}{*}{PS}}                                                                       & \multicolumn{3}{c|}{}                                                             & 81.72 & 94.75 \\
\multicolumn{1}{c|}{}                                                                                          & \multicolumn{3}{c|}{\checkmark}                                    & \textbf{90.12} & \textbf{96.73} \\ \hline
\multicolumn{1}{c|}{\multirow{5}{*}{\shortstack{Freezing\\Blocks}}} & 1-4                       & 5-8                       & 8-12                      &       &       \\ \cline{2-4}
\multicolumn{1}{c|}{}                                                                                          & \checkmark & \checkmark & \checkmark & 60.79 & 65.16 \\
\multicolumn{1}{c|}{}                                                                                          &                           & \checkmark & \checkmark & 82.17 & 90.16 \\
\multicolumn{1}{c|}{}                                                                                          &                           &                           & \checkmark & \textbf{90.12} & \textbf{96.73} \\
\multicolumn{1}{c|}{}                                                                                          &                           &                           &                           & 89.66 & 96.05 \\ \hline
\end{tabular}
\caption{\textbf{Ablation on Patch Shuffle (PS) and layer-freezing strategy.}
Detection accuracy (\%) and average precision (\%) on the GenImage dataset.}
\label{table:ablation2}
\end{table}

\section{Conclusion}
In this work, we investigate why Patch Shuffle exhibits unique effectiveness in CLIP-based AI-generated image detection. We show that PS suppresses global semantic continuity while preserving generator artifacts, thereby reducing semantic bias and homogenizing representations between natural and synthetic images. A layer-wise analysis further reveals that CLIP’s deep semantic structure acts as a regulator that stabilizes feature geometry and prevents overfitting. Leveraging these insights, we introduce SemAnti, a semantic-antagonistic fine-tuning paradigm that freezes the semantic subspace and adapts only the artifact-sensitive layers under shuffled semantics. SemAnti achieves state-of-the-art generalization on AIGCDetectBenchmark and GenImage, highlighting the importance of regulating semantics to unlock CLIP’s full potential for cross-domain artifact detection.
{
    \small
    \bibliographystyle{ieeenat_fullname}
    \bibliography{main}
}

\clearpage
\appendix
\setcounter{page}{1}
\maketitlesupplementary

This supplementary material provides additional results and analyses to support our main paper. Section~\ref{sec:cropping_effect} includes a controlled experiment demonstrating the necessity of using cropped inputs without any resizing in our preprocessing pipeline. Section~\ref{sec:extended_results} presents extended performance evaluations across benchmarks. Section~\ref{sec:additional_visuals} offers further visualization results to enhance the interpretability of our method. Section~\ref{sec:additional_ablation} reports additional ablation studies on model variants and different configurations.

\section{Effect of Cropping without Resizing in Preprocessing}
\label{sec:cropping_effect}
Here, we use Swin-T \cite{liu2021swin} trained on the SDv4 subset of the GenImage dataset. During evaluation, the model is tested on multiple unseen generative sources to assess cross source generalization. This experiment aims to examine whether the upsampling artifacts introduced during resizing become entangled with the original forgery traces in the image, thereby confusing the detector and reducing its ability to generalize. It further evaluates whether a cropping based strategy can better preserve the native artifact patterns that are essential for reliable AGID. We compare two preprocessing strategies. The crop strategy tiles the input image to exceed 224 $\times$ 224 resolution and then central crops it to 224 $\times$ 224, avoiding any interpolation. The resize strategy directly resizes the image to 224 $\times$ 224 using bilinear interpolation.

\begin{figure}[ht]
    \centering
    \includegraphics[width=\linewidth]{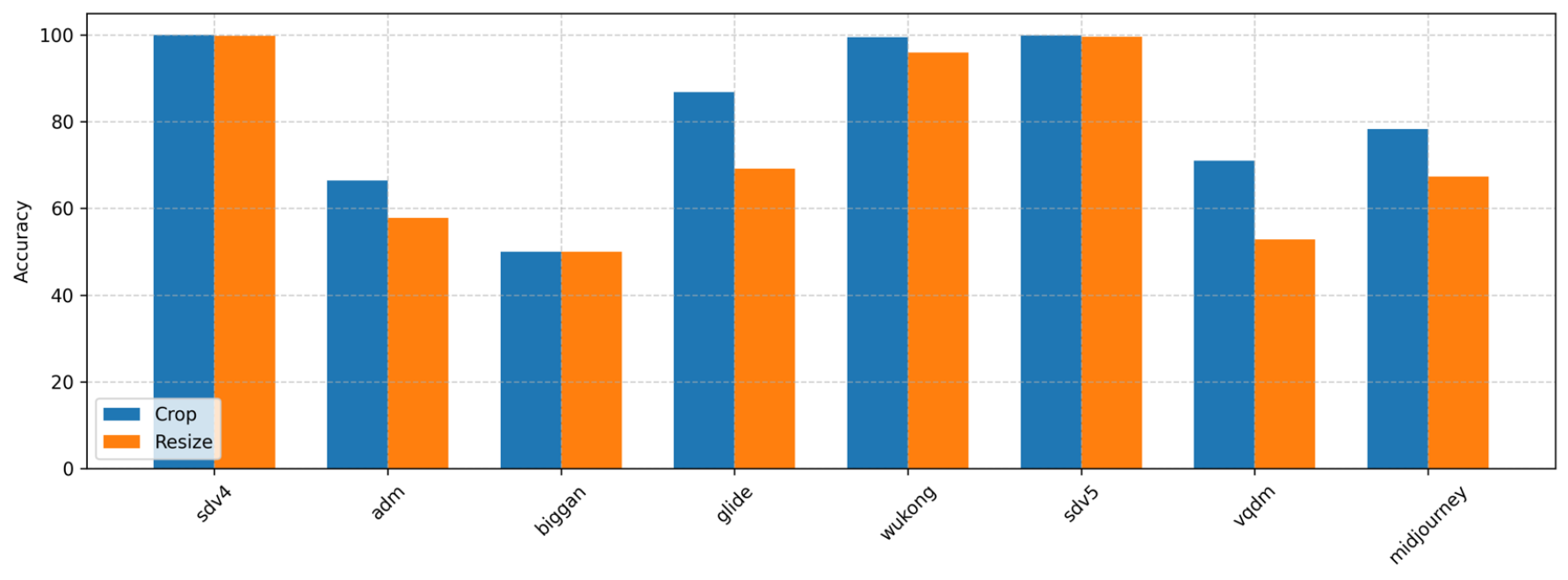}
    \caption{Comparison of performance (Accuracy \%) under two preprocessing strategies: cropping without resizing versus direct resizing. The model is trained on the SDv4 subset of GenImage and evaluated on all subsets.}
    \label{figure:app_bar}
\end{figure}

As shown in Figure~\ref{figure:app_bar}, across all evaluation subsets, cropping without resizing consistently yields higher or comparable accuracy relative to direct resizing. The improvement is substantial for ADM, GLIDE, VQDM, and Midjourney, indicating that interpolation based resizing disrupts the low level statistics and frequency signatures characteristic of generative models. These distortions lead to the mixing of artificial upsampling patterns with intrinsic synthetic artifacts, making the detector less capable of identifying genuine forgery cues. In contrast, cropping preserves the native pixel distribution and structural biases of the synthetic images, resulting in stronger generalization across diverse generators. The observed performance gains demonstrate the necessity of adopting a cropping based preprocessing pipeline when training AGI detectors.

\section{Extended Performance Results}
\label{sec:extended_results}
To complement the results reported in the main paper, we include additional metrics for both AIGCDetectBenchmark and GenImage. In the main text, we primarily presented AP on AIGCDetectBenchmark and Acc on GenImage. Here, we report the corresponding accuracy and AP results, respectively, to provide a complete evaluation under both metric conventions.

For AIGCDetectBenchmark, Table~\ref{table:appendix1} presents the accuracy scores obtained by training on the ProGAN subset and testing across sixteen generative sources. The accuracy trends closely match the AP results shown in the main paper. Our detector maintains strong performance on both GAN based and diffusion based generators, and the relative improvements over prior methods remain consistent across metrics. This alignment indicates that the model’s robustness is not tied to a specific evaluation measure.

For GenImage, Table~\ref{table:appendix2} provides the AP scores for all generative subsets. These results supplement the accuracy metrics reported in the main paper. The AP values exhibit the same relative ordering across methods and generator types, with our approach demonstrating stable performance gains on both in domain and out of domain subsets. The consistency between AP and accuracy further confirms that the learned representation transfers well across diverse synthesis pipelines.

Together, these additional results reinforce the conclusion that our method generalizes reliably across datasets, generators, and evaluation metrics, and that its improvements are not metric dependent but reflect genuinely more transferable forgery representations.
\begin{table*}[ht]
\resizebox{\linewidth
}{!}
{%
\begin{tabular}{lccccccccccccccccc}
\toprule
Method&\rot{ProGAN} &\rot{StyleGAN} & \rot{BigGAN} & \rot{CycleGAN} &\rot{StarGAN} &\rot{GauGAN} &\rot{StyleGAN2} &\rot{WFIR} &\rot{ADM} &\rot{Glide} & \rot{{Midjourney}} & \rot{{SD v1.4}} & \rot{{SD v1.5}}& \rot{{VQDM}}& \rot{{Wukong}}& \rot{{DALLE2}} & \rot{\textit{Mean}} \\ \midrule
CNNSpot \citep{wang2020cnn} & \textbf{100.00} & \textbf{99.83} & 85.99 & 94.94 & 99.04 & 90.82 & \textbf{99.48} & \textbf{99.85} & 75.67 & 72.28 & 66.24 & 61.20 & 61.56 & 68.83 & 57.34 & 53.51 & 80.41 \\
FreDect \citep{frank2020leveraging} & {\underline{99.99}} & 88.98 & 93.62 & 84.78 & 99.49 & 82.84 & 82.54 & 55.85 & 61.77 & 52.92 & 46.09 & 37.83 & 37.76 & 85.10 & 39.58 & 38.20 & 67.96 \\
Fusing \citep{ju2022fusing} & \textbf{100.00} & 85.20 & 77.40 & 87.00 & 97.00 & 77.00 & 83.30 & 66.80 & 49.00 & 57.20 & 52.20 & 51.00 & 51.40 & 55.10 & 51.70 & 52.80 & 68.38 \\
LNP \citep{liu2022detecting} & 99.67 & 91.75 & 77.75 & 84.10 & 99.92 & 75.39 & 94.64 & 70.85 & 84.73 & 80.52 & 65.55 & 85.55 & 85.67 & 74.46 & 82.06 & 88.75 & 83.84 \\
LGrad \citep{tan2023learning} & 99.83 & 91.08 & 85.62 & 86.94 & 99.27 & 78.46 & 85.32 & 55.70 & 67.15 & 66.11 & 65.35 & 63.02 & 63.67 & 72.99 & 59.55 & 65.45 & 75.34 \\
UnivFD \cite{ojha2023towards} & 99.81 & 84.93 & 95.08 & 98.33 & 95.75 & {\underline{99.47}} & 74.96 & 86.90 & 66.87 & 62.46 & 56.13 & 63.66 & 63.49 & 85.31 & 70.93 & 50.75 & 78.43 \\
DIRE \citep{wang2023dire} & 95.19 & 83.03 & 70.12 & 74.19 & 95.47 & 67.79 & 75.31 & 58.05 & 75.78 & 71.75 & 58.01 & 49.74 & 49.83 & 53.68 & 54.46 & 66.48 & 68.68 \\
NPR \citep{tan2024rethinking} & 99.79 & 97.70 & 84.35 & 96.10 & 99.35 & 82.50 & 98.38 & 65.80 & 69.69 & 78.36 & 77.85 & 78.63 & 78.89 & 78.13 & 76.11 & 64.90 & 82.91 \\
FatFormer \citep{liu2024forgery} & 99.89 & 97.13 & \textbf{99.50} & \textbf{99.36} & 99.75 & 99.43 & {\underline{98.80}} & 88.10 & 78.44 & 88.03 & 56.09 & 67.83 & 68.06 & 86.88 & 85.86 & 69.70 & 86.43 \\
DRCT \citep{drct} & 76.83 & 71.58 & 82.68 & 94.51 & 63.13 & 78.39 & 67.75 & 63.10 & 79.42 & 89.18 & \textbf{91.50} & 95.01 & 94.41 & 90.03 & {\underline{94.68}} & {\underline{92.55}} & 82.80 \\
D³ \citep{D3} & 99.89 & 90.84 & {\underline{97.84}} & 98.35 & 97.60 & \textbf{99.79} & 88.91 & 84.86 & 77.36 & 74.21 & 70.44 & 73.73 & 73.82 & 89.72 & 72.32 & 70.07 & 84.98 \\
CO-SPY \citep{co-spy} & 99.96 & 97.83 & 90.57 & {\underline{99.14}} & 94.06 & 93.29 & 98.62 & 70.26 & 75.17 & 86.44 & 87.23 & 85.77 & 85.73 & 81.02 & 80.95 & 75.84 & 87.62 \\
PatchCraft \citep{exper_RPTC} & \textbf{100.00} & 92.77 & 95.80 & 70.17 & {\underline{99.97}} & 71.58 & 89.55 & 85.80 & 82.17 & 83.79 & {\underline{90.12}} & {\underline{95.38}} & {\underline{95.30}} & 88.91 & 91.07 & \textbf{96.60} & 89.31 \\
AIDE \citep{yan2024sanity} & {\underline{99.99}} & {\underline{99.64}} & 83.95 & 98.48 & 99.91 & 73.25 & 98.00 & {\underline{94.20}} & \textbf{93.43} & \textbf{95.09} & 77.20 & 93.00 & 92.85 & {\underline{95.16}} & 93.55 & \textbf{96.60} & {\underline{92.77}}\\
\hline
Ours & \textbf{100.00} & 95.50 & 97.48 & \textbf{99.36} & \textbf{100.00} & 96.55 & 97.63 & 88.60 & {\underline{89.65}} & {\underline{93.80}} & 81.18 & \textbf{97.81} & \textbf{97.87} & \textbf{96.04} & \textbf{95.55} & 91.35 & \textbf{94.90} \\
\bottomrule
\end{tabular}
}
\caption{\textbf{Comparison on the AIGCDetectBenchmark \citep{exper_RPTC}}. Accuracy (\%) of different detectors (rows) in detecting natural and synthetic images from different generators (columns). These methods are trained on ProGAN and evaluated over sixteen generators. The best result and the second-best result are marked in \textbf{bold} and \underline{underline}, respectively.}
\label{table:appendix1}
\end{table*}
\begin{table*}[ht]
\centering
\resizebox{\linewidth}{!}
{\small
\begin{tabular}{lccccccccc}
\hline
Method & \multicolumn{1}{c}{Midjourney} & \multicolumn{1}{c}{SD v1.4} & \multicolumn{1}{c}{SD v1.5} & \multicolumn{1}{c}{ADM} & \multicolumn{1}{c}{GLIDE} & \multicolumn{1}{c}{Wukong} & \multicolumn{1}{c}{VQDM} & \multicolumn{1}{c}{BigGAN} & \multicolumn{1}{c}{\textit{Mean}} \\
\hline
CNNSpot \citep{wang2020cnn} & 73.53 & 99.91 & 99.87 & 47.93 & 72.70 & {\underline{99.76}} & 53.22 & 41.61 & 73.57 \\
FreDect \citep{frank2020leveraging} & 65.11 & \textbf{94.27} & \textbf{93.85} & 48.68 & 72.85 & \textbf{92.10} & 61.33 & 70.80 & 74.87 \\
Fusing \citep{ju2022fusing} & 77.94 & {\underline{99.98}} & {\underline{99.93}} & 55.17 & 77.29 & \textbf{99.92} & 61.18 & 45.01 & 77.05 \\
LNP \citep{liu2022detecting} & 74.55 & \textbf{99.88} & 99.84 & {\underline{86.69}} & {\underline{96.33}} & 99.60 & 88.51 & 42.67 & 86.01 \\
UnivFD \citep{ojha2023towards} & 78.08 & 96.58 & 96.43 & 53.77 & 81.92 & 87.38 & 65.56 & 67.07 & 81.25 \\
DIRE \citep{wang2023dire} & 63.94 & 97.51 & 97.53 & 68.60 & 85.61 & 95.66 & 64.55 & 44.76 & 77.27 \\
FreqNet \citep{tan2024frequency} & 67.55 & 94.18 & 94.39 & 78.21 & 92.47 & 91.64 & 75.60 & 40.56 & 79.32 \\
NPR \citep{tan2024rethinking} & 88.50 & 99.54 & 99.55 & 75.00 & \textbf{99.12} & 95.91 & 66.59 & 35.18 & 82.75 \\
DRCT \citep{drct} & 89.38 & 94.40 & 94.42 & 81.74 & \textbf{92.83} & 93.99 & {\underline{90.89}} & {\underline{74.58}} & 89.03 \\
CO-SPY \citep{co-spy} & {\underline{92.26}} & 96.92 & 96.95 & 81.63 & 95.90 & 96.72 & 90.57 & 65.39 & {\underline{89.54}} \\ \hline
\textbf{Ours} & \textbf{98.59} & \textbf{99.99} & \textbf{99.95} & \textbf{93.46} & 92.33 & 99.56 & \textbf{98.13} & \textbf{95.79} & \textbf{97.23} \\ \hline
\end{tabular}
}
\caption{\textbf{Comparison on the GenImage \citep{zhu2024genimage}}. Average precision (AP \%) of different detectors (rows) in detecting natural and synthetic images from different generators (columns). These methods are trained on 
natural images from ImageNet and synthetic images generated by SD v1.4 and evaluated over eight generators. The best result and the second-best result are marked in \textbf{bold} and \underline{underline}, respectively.}
\label{table:appendix2}
\end{table*}

\section{Additional Visualizations}
\label{sec:additional_visuals}
In this section, we further provide t-SNE visualizations of SemAnti on six additional generators, including BIGGAN, GLIDE, VQDM, SDv5, Midjourney, and Wukong, under both settings without PS and with PS. These plots complement the main paper and examine whether the representation patterns we observe on SDv4 and ADM generalize to a broader set of generation models.

\begin{figure*}[ht]
    \centering
    \includegraphics[width=0.97\linewidth]{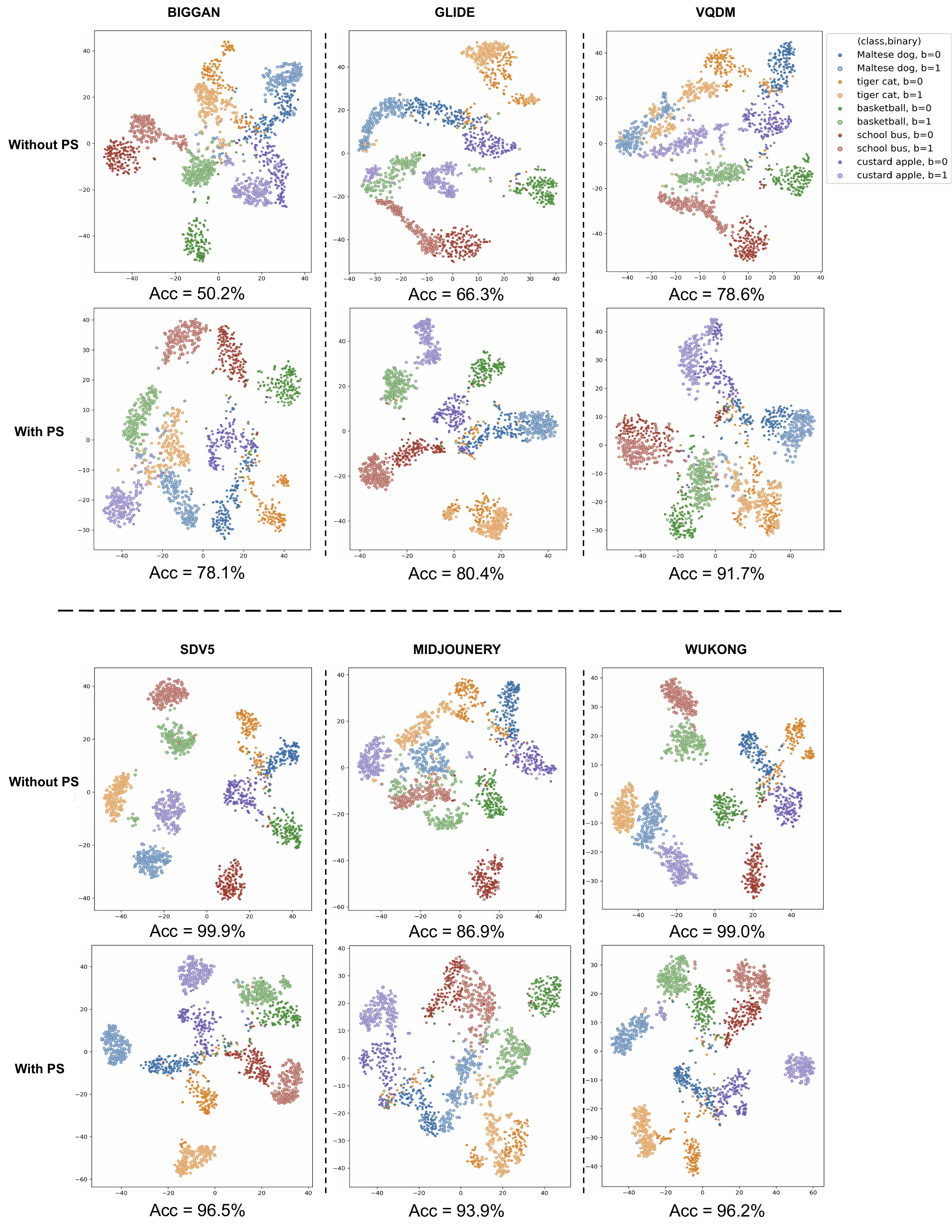}
    \caption{More visualizations of the latent distributions of our method under different settings.}
    \label{figure:appendix_visual}
\end{figure*}

As illustrated in Figure~\ref{figure:appendix_visual}, without PS, the behavior is highly distribution dependent. On SDv5 and Wukong, whose distributions are close to the training set SDv4, the latent space remains clean and well structured, and the model reaches almost perfect accuracy. The t-SNE plots show clear and compact clusters and an apparently stable separation between natural and synthetic samples, which superficially suggests that the model has learned a good representation. However, on BIGGAN, GLIDE, and VQDM, where the distributions deviate more from the training domain, the structure deteriorates quickly. Many semantic clusters become heavily mixed, category boundaries blur, and natural and synthetic samples become entangled, with no obvious global organization in the embedding space. This mirrors what we observe on the OOD ADM set in the main paper, where the model trained without PS collapses into an entangled representation that fails to maintain a reliable natural–synthetic boundary and consequently suffers from weak generalization despite high in-domain accuracy.

After introducing PS, the representations become markedly more stable and consistent across all generators. On all six test sets, the t-SNE visualizations exhibit a similar organized pattern to that reported in the main paper. Semantic neighbors remain close in the latent space, which indicates that the model preserves meaningful semantic structure. At the same time, natural and synthetic samples are arranged along a more coherent authenticity direction. Natural images tend to concentrate near the centers of semantic clusters, while synthetic images are distributed toward the periphery or along a consistent direction around each cluster. This two-dimensional organization, in which one axis reflects semantics and the other reflects authenticity, appears not only on SDv5 and Wukong but also on BIGGAN, GLIDE, and VQDM, which are more challenging and distributionally distant from SDv4. Correspondingly, the model with PS achieves substantially higher accuracies on these OOD generators, for example improving from 50.2\% to 78.1\% on BIGGAN, from 66.3\% to 80.4\% on GLIDE, and from 78.6\% to 91.7\% on VQDM, while only slightly reducing performance on SDv5 and Wukong compared to the overfitted no-PS models.

These additional visualizations reinforce the positive interpretation given in the main paper. When CLIP is trained directly without PS, the feature space is dominated by global semantics. The model tends to memorize semantic artifacts correlations between object categories. As a result, it often achieves near-perfect separation of natural and synthetic images on the training distribution but only by distorting the semantic geometry and breaking semantic coupling. Once evaluated on unseen generators, this brittle structure collapses, and both semantic clusters and the natural–synthetic boundary become unstable. With PS, we explicitly weaken CLIP’s reliance on global semantic patterns and force the model to rely on semantic-consistent artifact cues instead. The learned embedding preserves local semantic neighborhoods and simultaneously maintains a robust authenticity axis that aligns natural samples and synthetic samples in a structured way. This allows SemAnti to retain strong in-domain performance while significantly improving robustness and interpretability on OOD generators.

\section{Additional Ablation Studies}
\label{sec:additional_ablation}
In the main paper, we study the impact of shuffle granularity by varying the patch size from 8 to 32 on two CLIP backbones, ViT-L/14 and ViT-B/32, and we report mean accuracy over AIGCDetectBenchmark and GenImage. The results show that the optimal performance is achieved when the shuffle size is set to 16 for ViT-L/14 and 14 for ViT-B/32, which we attribute to a balance between weakening global semantics and preserving local artifact consistency. Smaller patches do not sufficiently disrupt semantic correlations, while overly large patches destroy fine-grained artifacts and lead to degraded generalization.

Table~\ref{table:appendix_ablation2} in the appendix provides a more comprehensive view of this effect. Besides ViT-L/14 and ViT-B/32, we additionally evaluate a ViT-B/16 variant, and we report not only overall accuracy but also average precision (AP) on both AIGCBenchmark and GenImage. Across all three backbones, the same trend consistently appears. For ViT-L/14, a patch size of 16 yields the highest accuracy and AP on AIGCBenchmark while maintaining competitive performance on GenImage, whereas a size of 32 leads to a clear drop in accuracy despite relatively stable AP. ViT-B/16 behaves similarly, with medium patch sizes preserving both high accuracy and strong ranking quality, and with performance deteriorating once the shuffle blocks become too large. ViT-B/32 again peaks at a patch size of 14, with both smaller and larger sizes causing noticeable declines, especially in accuracy.

The joint accuracy and AP results support our interpretation that patch shuffle acts as a controllable perturbation on CLIP’s semantic structure. When the shuffle size is too large, the global layout of objects and scenes remains largely intact, which allows the model to continue exploiting semantic shortcuts and therefore limits the benefit of PS for OOD generalization. When the shuffle size is too small, semantic information is heavily fragmented and, at the same time, many subtle generator artifacts are also disrupted, which harms both detection accuracy and the quality of the ranking measured by AP. In contrast, moderate shuffle sizes around 14–16 induce a regime in which global semantic configurations are sufficiently homogenized across domains while local texture and artifact patterns remain distinguishable. This regime leads to the most robust cross-generator performance and aligns well with the t-SNE analyses in the main text, where PS encourages a representation space that preserves semantic neighborhoods while organizing natural and synthetic images along a separate authenticity axis.
\begin{table}[ht]
\resizebox{\linewidth
}{!}
{%
\begin{tabular}{cccccc}
\hline
\multirow{2}{*}{Variant} & \multirow{2}{*}{Patch Size} & \multicolumn{2}{c}{AIGCBenchmark} & \multicolumn{2}{c}{Genimage} \\ \cline{3-6} 
 &  & Acc & AP & Acc & AP \\ \hline
\multirow{4}{*}{ViT-L/14} & 8 & 91.43 & 97.46 & 71.85 & 91.68 \\
 & 14 & 90.18 & 97.82 & \textbf{86.08} & \textbf{98.16} \\
 & 16 & \textbf{94.90} & \textbf{99.03} & 73.28 & 90.89 \\
 & 32 & 80.88 & 96.92 & 80.19 & 97.82 \\ \hline
\multirow{4}{*}{ViT-B/16} & 8 & 89.60 & 98.11 & \textbf{87.13} & \textbf{97.15} \\
 & 14 & \textbf{91.43} & \textbf{98.32} & 84.21 & 92.89 \\
 & 16 & 84.60 & 92.32 & 82.29 & 94.83 \\
 & 32 & 84.70 & 95.94 & 86.29 & 95.47 \\ \hline
\multirow{4}{*}{ViT-B/32} & 8 & 76.78 & 88.92 & 83.87 & 92.82 \\
 & 14 & 81.81 & 88.79 & \textbf{91.32} & \textbf{97.23} \\
 & 16 & 71.22 & 92.07 & 84.15 & 94.71 \\
 & 32 & \textbf{83.99} & \textbf{93.20} & 84.05 & 97.57 \\ \hline
\end{tabular}
}
\caption{Extended analysis of patch shuffle size. Detection accuracy (Acc, \%) and average precision (AP, \%) on AIGCBenchmark and GenImage for three CLIP-based variants under different shuffle granularities.}
\label{table:appendix_ablation2}
\end{table}

\end{document}